\documentclass[smallabstract,smallcaptions]{dccpaper}
\usepackage{epsfig}
\usepackage{citesort}
\usepackage{wrapfig}
\usepackage{amsmath}
\usepackage{amssymb}
\usepackage{color}
\usepackage{url}
\usepackage{multirow}
\usepackage{booktabs}
\usepackage{subfigure}
\newlength{\figurewidth}
\newlength{\smallfigurewidth}
\usepackage{bbding} 
\begin{document}

\title
{\large
\textbf{AWEQ: Post-Training Quantization with Activation-Weight Equalization for Large Language Models}
}

\author{%
Baisong Li$^{1, 2}$, Xingwang Wang$^{\star, 1, 2}$, Haixiao Xu$^{1, 2}$\\[0.5em]
\thanks{$^\star$Corresponding author}
{\small\begin{minipage}{\linewidth}\begin{center}
\begin{tabular}{ccc}
$^{1}$ School of Computer Science and Technology, Jilin University  \\
$^{2}$ Key Laboratory of Symbolic Computation and Knowledge Engineering \\ of Ministry of Education, Jilin University   \\ 
\texttt{lbs23@mails.jlu.edu.cn} \quad
\texttt{\{xww, haixiao\}@jlu.edu.cn}
\end{tabular}
\end{center}\end{minipage}}
}
\maketitle
\thispagestyle{empty}
\begin{abstract}
Large language models(LLMs) excellent performance across a variety of tasks, but they come with significant computational and storage costs. Quantizing these models is an effective way to alleviate this issue. However, existing methods struggle to strike a balance between model accuracy and hardware efficiency. This is where we introduce AWEQ, a post-training method that requires no additional training overhead. AWEQ excels in both ultra-low-bit quantization and 8-bit weight and activation (W8A8) quantization. There is an observation that weight quantization is less challenging than activation quantization. AWEQ transfers the difficulty of activation quantization to weights using channel equalization, achieving a balance between the quantization difficulties of both, and thereby maximizing performance. We have further refined the equalization method to mitigate quantization bias error, ensuring the robustness of the model. Extensive experiments on popular models such as LLaMA and OPT demonstrate that AWEQ outperforms all existing post-training quantization methods for large models.

\end{abstract}
\section{Introduction}
LLMs have demonstrated outstanding performance across various tasks. However, due to the immense model size and computational overhead, it is challenging to run and deploy them on resource-constrained computing devices. Quantization is a promising approach to reduce the storage and computational costs of large models by representing model parameters with smaller bit representations. Quantization can be categorized into Quantization-Aware Training (QAT) and Post-Training Quantization (PTQ). QAT simulates the quantization effect during the training process, requiring the addition of quantization-aware modules to the model and restarting to train, which is often unacceptable for LLMs. PTQ, on the other hand, only involves parameter reconfiguration of a trained model and does not incur additional training overhead.


Recent works\cite{obc,yao2022zeroquant,frantar-gptq}, such as GPTQ\cite{frantar-gptq}, provide a quantitative analysis of the effects of quantizing individual weight values on model performance. GPTQ addresses the challenge of compensating for unquantized weights using second-order information, specifically the inverse hessian matrix. This approach relies on reordering techniques for the quantization of specific models, such as OPT-66B and LLaMA-7B. It's important to note that updating the inverse hessian matrix can be hardware-intensive, resulting in slower quantization speeds. On the other hand, GPT-Q aims to expedite quantization by exclusively considering the influence of weights within each row during implementation, although this may lead to a partial decrease in accuracy.

Based on the observation that the quantization difficulty of weights is lower than that of activations, SmoothQuant\cite{smoothquant} and AWQ\cite{lin2023awq} define the quantization difficulty based on the maximum absolute values of activations and weights, and then shifts the difficulty of activation quantization to weights to improve performance. However, our experiments have shown that representing the quantization difficulty using the ratio of the per-channel range in activations and weights to the range of the tensor can lead to more performance improvements. Furthermore, quantization combined with equalization introduces a biased quantization error[7]. DFQ corrects the biased quantization error by utilizing statistics from the Batch Normalization (BN) layer. 

Based on these above observations, in this paper, we propose Activation-Weight Equalization Quantization (AWEQ), using the ratio of the per-channel range in activations and weights to define the quantization difficulty, leveraging the dynamic statistical quantization bias error correction to robust the quantized model. 
Extensive experiments were conducted on widely-used LLaMA\cite{touvron2302llama} and OPT\cite{zhang2022opt} models. Our approach was evaluated in both ultra-low-bit quantization scenarios and the W8A8 scenario, achieving state-of-the-art results. This demonstrates that AWEQ surpasses all existing post-training quantization methods for LLMs.
\section{Background}
\begin{figure} 
\centering 
\subfigure[\textbf{Before AWE}, the weight distribution is \underline{very easy to quantize}.
]{\label{fig:subfig:w_b}
\includegraphics[width=0.45\linewidth]{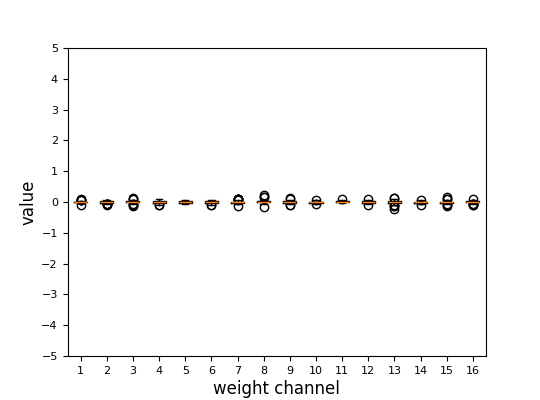}}
\hspace{0.01\linewidth}
\subfigure[\textbf{After AWE}, the weight distribution is \underline{easy to quantize}.
]{\label{fig:subfig:w_a}
\includegraphics[width=0.45\linewidth]{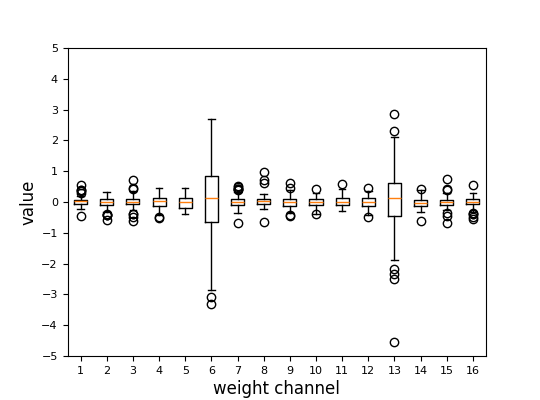}}
\vfill
\subfigure[\textbf{Before AWE}, the distribution of activation values is \underline{hard to quantize}.
]{\label{fig:subfig:a_b}
\includegraphics[width=0.45\linewidth]{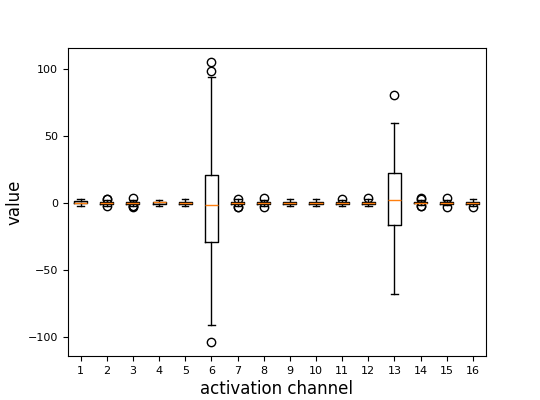}}
\hspace{0.01\linewidth}
\subfigure[\textbf{After AWE}, the distribution of activation values is \underline{easy to quantize}.
]{\label{fig:subfig:a_a}
\includegraphics[width=0.45\linewidth]{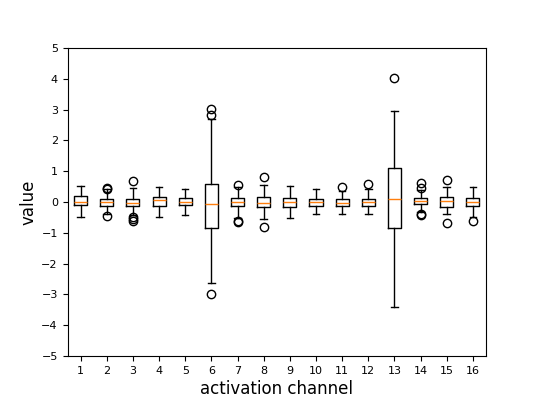}}
\caption{Following the approach of SmoothQuant, we examine the data distribution of the 15th decoder (with other layers showing similar distributions) in the LLaMA-7B model before and after the execution of the AWE operation. We randomly select 16 channels to illustrate the distribution using box plots. In each subplot, the horizontal axis represents the channels of weights or activations, and the vertical axis represents the values of activations or weights. Quantization difficulty level is indicated using \underline{underline}.}
\label{fig:subfig}

\end{figure}

\textbf{Quantization.} Quantization is a method of mapping values from high precision to low bits. Large language models are often saved as FP16 values, and mapping them to lower bits (especially INT8) will lead to speedup in inference and reduced storage overhead. The quantification formula is as follows:
\begin{equation}
    x_{q}=clamp\left(\lfloor\frac {x}{Step}\rceil+Z;q_{min},q_{max}\right)
\end{equation}
\begin{equation}
    Step=\left(x_{max}-x_{min}\right)/\left(2^b-1\right)
\end{equation}
Here, $x$ represents the high-precision tensor value, and $x_q$ represents the quantized value. $\left\lfloor\cdot\right\rceil$ represents the round-to-nearest operator. $Step$ signifies the quantization scaling factors, while $Z$ denotes the offset defined as the zero-point. $x_{max}$ stands for the maximum value in the vector, and $x_{min}$ represents the minimum value in the vector. The quantization range, denoted as $[q_{min}, q_{max}]$, is determined by the bit-width which denoted as $b$. For simplicity, we are exclusively considering uniform unsigned symmetric quantization for hardware friendly.\\ 
\textbf{Quantization Difficulty.} According the observation of SmoothQuant, weight quantization is generally less challenging than activation quantization. Neural network weights typically follow a normal distribution(as shown in Figure \ref{fig:subfig:w_b}), while some channels in neural network activations often exhibit outliers(as shown in Figure \ref{fig:subfig:a_b}). The presence of outliers results in a significant waste of quantization grid points, leading to a substantial increase in noise introduced by quantization.

Previous works\cite{obc,frantar-gptq}, such as GPTQ\cite{frantar-gptq}, often focuse solely on the quantization of the weights themselves, without considering the significant impact of the distribution of activation values on quantization. GPTQ employe a complex and hardware-unfriendly method to quantitatively analyze weight errors induced by quantization and compensate for these quantization errors. This approach is highly impractical for increasingly large language models. LLM.int8()\cite{llmint8} is a method that separates outliers from normal values. Outliers are not included in the quantization process. LLM.int8() compared to the original 8-bit quantization, yields significant improvements. However, this method requires manual specification of the outlier detection criteria and is a per-channel mixed-precision quantization, which is not hardware-friendly.

SmoothQuant defines the quantization challenge by considering the maximum absolute values of activations and weights. It then transfers the difficulty of quantizing activations to the weights to enhance performance. Nevertheless, our experiments have revealed that representing the quantization challenge as the ratio of the per-channel range in activations and weights to the tensor's range can yield even greater performance improvements.
\begin{figure}[!ht]
\centering
\includegraphics[scale=0.143]{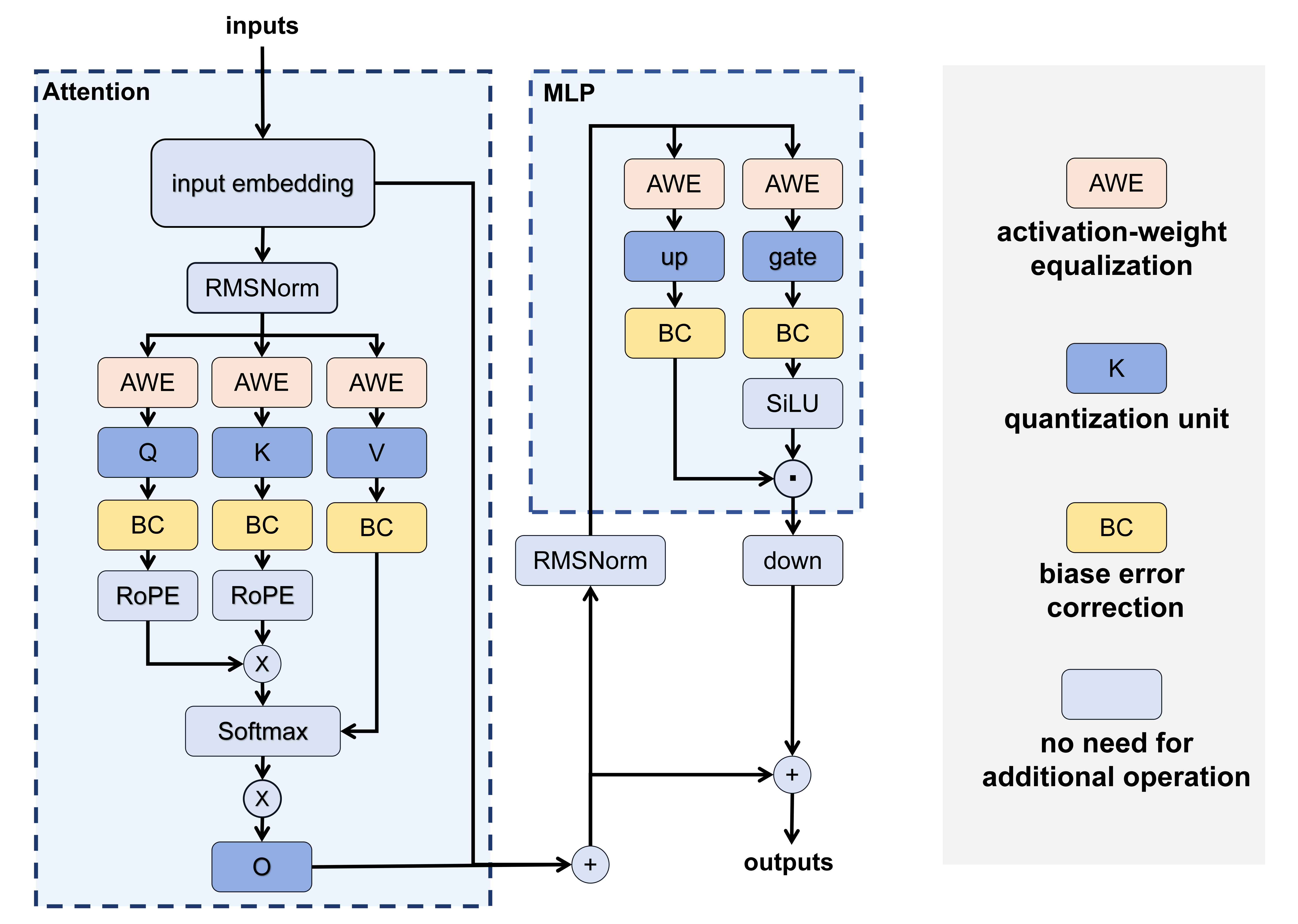}
\caption{Overview of our work in \textbf{LLaMA Decoder}. LLMs are typically constructed by stacking \textit{transformer} decoders or encoders (e.g., LLaMA-7B consists of 32 stacked decoders). \textbf{Attention} denotes the decoder's self-attention module, and \textbf{MLP} represents the decoder's multi-layer perceptron module. Here, we illustrate our approach for quantizing the decoder in LLaMA.}
\end{figure}
\label{overview}
\section{Methodology}
Per-channel quantization methods often require models to have a high throughput to ensure a certain scale of values that are being quantized, thus guaranteeing the effectiveness of quantization. This is impractical for LLMs. AWEQ  exclusively focuses on per-tensor quantization, which is both efficient and hardware-friendly. The AWE operation takes place in the stage preceding quantization, with the aim of simultaneously equalizing activations and weights on each channel to ensure that both weights and activations have favorable value distributions. This equalization and quantization process introduces a bias error, which we address with the proposed quantization Bias Correction (BC) method. An overview of AWEQ is shown in Figure \ref{overview}.
\subsection{Preliminaries}
To shift the primary quantization challenges from activations to weights and equalize the weight values and activations to the same range, SmoothQuant employed a per-channel equalization method, mathematically represented as follows:
\begin{equation}
    \mathbf{Y}=(\mathbf{X}\text{diag}(\mathbf{s})^{-1})\cdot(\text{diag}(\mathbf{s})\mathbf{W})=\hat{\mathbf{X}}\cdot\hat{\mathbf{W}}
    \label{smooth_eq}
\end{equation}
Where $S = \text{diag}(s)$ is a diagonal matrix, with $S_{ii}$ representing the equalization factor $s_i$ for channel $i$. $\hat{\mathbf{X}}$ and $\hat{\mathbf{W}}$ represent the equalized activations and weights and $\hat{\mathbf{X}}=\mathbf{X}\text{diag}(\mathbf{s})^{-1}$, $\hat{\mathbf{W}}=\text{diag}(\mathbf{s})\mathbf{W}$. For LLMs, this equalization operation typically occurs at the entrances of self-attention blocks and fully connected layers. Quantization of activation values can often be seamlessly fused with the previous block. At the inference stage, it will not introduce additional computational overhead while maintaining the model's throughput.
\subsection{Quantization with Activation-Weight Equalization}
The essence of our method lies in adjusting the weight range to map out-of-range activation values(i,e., outliers) into an appropriate range, ultimately resulting in reduced quantization error. When the range of each channel's activations and weights equals the overall tensor's range, it indicates that we are using the best representation for each channel. SmoothQuant defines the quantization challenge by considering the maximum absolute values of activations and weights. DFQ\cite{DFQ} uses the product of the ratio of the channel range of adjacent two layers' weights to the tensor range to equalize the weights of adjacent two layers. Inspired by DFQ, we here defines the quantization challenge as the ratio of the per-channel range in activations and weights to the tensor's range:
\begin{equation}
    \mathbf{p}_i^{(\hat{W})}=\frac{\mathbf{r}_i^{(\hat{X})}}{R^{(\hat{X})}} ,\quad \mathbf{p}_i^{(\hat{X})}=\frac{\mathbf{r}_i^{(\hat{W})}}{R^{(\hat{W})}} 
\end{equation}
 $\hat{W}$ and $\hat{X}$ represent the equalized $W$ and $X$ as shown in Equation \ref{smooth_eq}. $\mathbf{r}_i^{(\hat{X})}$ and  $\mathbf{r}_i^{(\hat{W})}$ respectively represent the range of the equalized activation and weight for the i-th channel, specifically, it is the maximum value minus the minimum value for each channel. ${R^{(\hat{X})}}$ and ${R^{(\hat{W})}}$ correspondingly denote the range of the equalized activation tensor and weight tensor, specifically, it is the maximum value minus the minimum value for all channel. We aim to find \(s\) such that the total precision per channel is maximized:
\begin{equation}
    \mathbf{s} = \mathop{\arg\max}_{\mathbf{s}}\sum_i\mathbf{p}_i^{(\hat{X})}\mathbf{p}_i^{(\hat{W})}
\end{equation}
 The optimization objective is transformed into:
\begin{align}
&\mathop{\arg\max}_\mathbf{s}\sum_i\mathbf{p}_i^{(\hat{X})}\mathbf{p}_i^{(\hat{W})}=\mathop{\arg\max}_\mathbf{s}\sum_i\frac{\mathbf{r}_i^{(\hat{X})}\mathbf{r}_i^{(\hat{W})}}{R^{(\hat{X})}R^{(\hat{W})}} \\
&=\sum_i\mathbf{r}_i^{(X)}\mathbf{r}_i^{(W)}\mathop{\arg\max}_\mathbf{s}\frac1{\max_j(\frac1{\mathbf{s}_j}\mathbf{r}_j^{(X)})\cdot\max_k(\mathbf{s}_k\mathbf{r}_k^{(W)})}
\label{eq:optim_range}
\end{align}
$\mathbf{r}_i^{(X)}$ and $\mathbf{r}_i^{(W)}$ represent the range of activation and weight for the $i$-th channel without equalization. $\max_j\left(\frac{1}{\mathbf{s}_j}\mathbf{r}_j^{(X)}\right)$, $\max_k\left(\frac{1}{\mathbf{s}_k}\mathbf{r}_k^{(W)}\right)$ obtain the maximum values of the equalized weights and activations channel-wise, ultimately yielding the maximum values of the tensor X and W. Next, we only need to minimize the denominator in Equation \ref{eq:optim_range} as:
\begin{equation}
   \mathbf{s} = \mathop{\arg\min}_\mathbf{s}\left(\max_j\left(\frac{1}{\mathbf{s}_j}\mathbf{r}_j^{(X)}\right)\cdot\max_k(\mathbf{s}_k\mathbf{r}_k^{(W)})\right)
    \label{mini_s_final}
\end{equation}
Assuming $\arg\max_j\frac{1}{\mathbf{s}_j}\mathbf{r}_j^{(X)}\neq \arg\max_k\mathbf{s}_k\mathbf{r}_k^{(W)}$, where $q$ represents the channel for which the product $\mathbf{s}_q\mathbf{r}_q^{(W)}$ is the second largest (while $q$ is not equal to $j$), we can find numerous solutions that satisfy Equation \ref{mini_s_final}. In these solutions, the new weight equalization factor of channel $k$, i.e, $s_k^{\prime}$ all fall within the range $[s_q, s_k]$. To ensure the uniqueness and robustness of Equation \ref{mini_s_final}, we directly set:
\begin{equation}
    \arg\max_j\frac{1}{\mathbf{s}_j}\mathbf{r}_j^{(X)}=\arg\max_k\mathbf{s}_k\mathbf{r}_k^{(W)}
    \label{constraint}
\end{equation}
To further address the issue of diversity in the solutions for $s$, we directly assume $ \mathbf{r}_{i}^{(\hat{X})}=\mathbf{r}_{i}^{(\hat{W})}$ for all channels. In the case of weight distributions in LLMs, this setting is generally equivalent to Equation \ref{constraint}. The equalization factor can ultimately be expressed as:
\begin{equation}
    \mathbf{s}_i=\frac{1}{\mathbf{r}_i^{(W)}}\sqrt{\mathbf{r}_i^{(X)}\mathbf{r}_i^{(W)}}
\end{equation}
\subsection{Quantization Bias Correction}
Quantization operations introduce bias errors in the output of neural networks, particularly in LLMs, where this biased quantization error accumulates as the neural network deepens. In this section, we demonstrate how to correct the bias error introduced by quantization in large models to ensure their robustness.
Due to the absence of Batch Normalization layers in most LLMs as DFQ mentioned,
we leverage the dynamic statistical quantization bias error correction to robust the quantized model: 
\begin{equation}
    \mathbb{E}[\widetilde{\mathbf{y}}_j - \mathbf{y}_j] \approx \frac{1}{N} \sum_n ((\widetilde{\mathbf{W}}\mathbf{x}_n)_j - (\mathbf{W}\mathbf{x}_n)_j)
    \label{bias_error}
\end{equation} where $y_{j}$ and $\widetilde{y}_{j}$ represent the output of network block $j$ before and after quantization, and $N$ represents the number of input data used to count the quantization error.
We define the distance between weight quantization before and after as $\boldsymbol{\epsilon} = \widetilde{\mathbf{W}} - \mathbf{W}$. Equation \ref{bias_error} can be represented as:
\begin{align}
    \mathbb{E}[\widetilde{\mathbf{y}} - \mathbf{y}]
    &\approx \mathbb{E}[\widetilde{\mathbf{y}}]- \mathbb{E}[\mathbf{y}] \\
    &\approx \frac{1}{N} \sum_n \left((\mathbf{W} + \boldsymbol{\epsilon}) \mathbf{x}_n - (\mathbf{W}\mathbf{x}_n)\right) \\
    &\approx \frac{1}{N} \sum_n (\boldsymbol{\epsilon}\mathbf{x}_n) \\ 
    &\approx \epsilon \mathbb{E}[x]
\end{align}

The corrected output, denoted as $\tilde{\mathbf{y}_j}$, is obtained by subtracting $\epsilon\mathbb{E}[x]$ from $\widetilde{\mathbf{y}}_j$. 
The expectation of the network block $j$ activations, $\mathbb{E}[x]$, will be statistically estimated alongside the normalization coefficient $S$ before quantization. This will not introduce additional computational and storage overhead during inference in LLMs.

\section{Experiments}
\subsection{Datasets and Bechmarks}
We conducted several zero-shot(i.e., no prompts were provided before the test) evaluation tasks: \textbf{PIQA (Everyday Physics Questions)}\cite{bisk2020piqa}: PIQA assesses large models' ability to understand everyday physics with over 16,000 training question-answer pairs. \textbf{HellaSwag (Commonsense Reasoning Dataset)}\cite{zellers2019hellaswag}: HellaSwag challenges language models with 70,000 multiple-choice questions that test their ability to extract information and summarize it. Human accuracy is 95.6\%. \textbf{WinoGrande (Large-Scale Question Dataset)}\cite{sakaguchi2021winogrande}: WinoGrande, is a sizable dataset with 44,000 questions designed to be more extensive and challenging. \textbf{ARC-e (Reasoning Challenge Dataset)}\cite{arc}: ARC is a multiple-choice question-answering dataset covering science questions from grades three to nine, with an easy version(i.e., ARC-e). Most questions have four answer choices, some with three or five. In all four tasks, the evaluation metric used is accuracy.

\subsection{Baslines}
In ultra-low-bit quantizations, we employed RTN(round-to-nearest) and GPTQ\cite{frantar-gptq} as baselines. For W8A8, we used ZeroQuant\cite{yao2022zeroquant}, LLM.int8()\cite{llmint8} and SmoothQuant\cite{smoothquant} as baselines. We separately tested and compared the accuracy of the baseline on the aforementioned four tasks and their average.
\subsection{Implementation}
All experiments were conducted on A800 GPUs, each with 80GB of VRAM. We implemented AWEQ using PyTorch.  We conducted experiments on the LLaMA\cite{touvron2302llama} and OPT\cite{zhang2022opt} models. We used the lm-eval-harness\footnote{https://github.com/EleutherAI/lm-evaluation-harness} to assess the model performance, this project provides a unified framework to test generative language models on a large number of different evaluation tasks. 
\subsection{Results}
\textbf{Ultra-low-bit quantization results.} 
We initially focused our research on the LLaMA model, compared to other open-source LLMs, it exhibits superior performance. It also serves as the foundation for many popular open-source models. To validate the effectiveness of our model, we begin by conducting ultra-low-bit weight quantization on the LLaMA model. This ultra-low-bit weight quantization for the LLaMA model ensures that a significant portion of the model's performance is preserved, which is highly advantageous in practical applications. As shown in Table \ref{tab:zero-task-eval}, in the ultra-low-bit quantization of weights, we consistently outperform RTN and GPTQ in the PIQA, HellaSwag, and WinoGrande tasks. For the average accuracy across the four tasks, our accuracy also surpasses that of RTN and GPTQ.

We also assess the effectiveness of AWEQ on LLaMA models with varying parameter sizes. The accuracy in Table \ref{tab:llama-model-eval} represents the average accuracy of the model across the four mentioned tasks. The results demonstrate that our model outperforms RTN and GPTQ on LLaMA models with different parameter size.
\begin{table}[!ht]
\caption{In the LLaMA-7B model, our focus was on low-bit quantization of weights while retaining activation precision. We tested the accuracy of the quantized models on four different tasks and provided their averages, \textbf{Avg }in table represents the average accuracy across the four tasks. The best results are indicated in bold.}
\label{tab:zero-task-eval}
 \centering
\begin{tabular}{ccccccc}
  \toprule
                     \textbf{LLaMA-7B}& & PIQA    & HellaSwag  & WinoGrande   & ARC-e   & Avg   \\ \midrule
FP16                  &-& 78.36\%          & 56.44\%          & 67.12\%          & 67.31\% & 67.31\% \\ \midrule
\multirow{3}{*}{INT3} &RTN&75.77\%          & 53.02\%          & 63.32\%          & \textbf{65.91\%} & 64.51\% \\
                      &GPTQ &70.92\%          & 46.74\%          & 60.89\%          & 60.10\%          & 59.66\% \\
                      &AWEQ&\textbf{76.46\%} &\textbf{ 54.11\%} & \textbf{65.89\% }& 65.86\%          & \textbf{65.58\%} \\
                      \midrule
\multirow{3}{*}{INT4} &RTN&77.90\%          & 55.83\%          & 65.62\%          & \textbf{66.29\% }    &66.41\% \\
                      &GPTQ&77.22\%          & 53.96\%          & 65.65\%          & 61.61\%              & 64.61\% \\
                      & AWEQ&\textbf{78.12\%} &\textbf{55.88\% } & \textbf{65.98\%} & 66.27\%              & \textbf{66.56\%} \\ \bottomrule
\end{tabular}
\end{table}

\textbf{INT8 quantization results.} 
By equalizing activations and weights, AWEQ can quantize models with activations that are more challenging to quantization. To demonstrate the generality and effectiveness of our approach, here we evaluate the performance of INT8 quantization(i.e., W8A8) on OPT-175B after completing the quantization, the results as shown in Table \ref{tab:opt-175-w8a8}. The results indicate that under the W8A8 quantization setting, AWEQ achieved state-of-the-art (SOTA) performance on HellaSwag, WinoGrande, and ARC-e tasks. The average performance achieved the best results in four tasks.

\begin{table}[!ht]
\caption{We tested the\textbf{ average accuracy }across the above four tasks of the LLaMA family, with the best experimental results highlighted in bold.}
\label{tab:llama-model-eval}
\centering
\begin{tabular}{ccccc}
 \toprule
\textbf{LLaMA} & 7B               & 13B              & 30B               & 65B             \\ \midrule
FP16          & 67.28\%          & 70.62\%          & 73.01\%           & 74.52\%          \\ \midrule
RTN           & 64.52\%          & 68.62\%          & 72.10\%            & 72.60\%          \\
GPTQ          & 59.65\%          & 68.69\%          & 70.8\%            & 73.13\%         \\
AWEQ          & \textbf{65.55\%} & \textbf{69.10\%} & \textbf{72.15\%}  & \textbf{73.34\%}\\ \bottomrule
\end{tabular}
\end{table}
\begin{table}[!ht]
\caption{In the W8A8 quantization setting, we tested the accuracy across above four tasks of OPT-175B, with the best results highlighted in bold. The result $^\star$ is obtained using the open-source code of SmoothQuant.}
\label{tab:opt-175-w8a8}
 \centering
\begin{tabular}{ccccccc}
  \toprule
                     \textbf{OPT-175B} & PIQA    & HellaSwag  & WinoGrande   & ARC-e   & Avg    \\ \midrule
    FP16                  & 79.76\%          & 59.37\%          & 72.68\%          & 71.10\% & 70.73\% \\ \midrule
    W8A8                  & 53.43\%          & 25.61\%          & 50.32\%          & 48.26\% & 44.40\% \\
    ZeroQuant             & 51.72\%          & 26.07\%          & 49.29\%          & 47.92\% & 43.75\% \\
    LLM.int8()            &\textbf{ 79.74\%  }        & 59.25\%          & 72.11\%          & 70.11\% & 70.30\% \\
    SmoothQuant            & 79.70\%          & 59.20\%          & 71.20\%          & 70.18\%$^\star$ & 70.07\% \\
    AWEQ                  & 79.72\%          & \textbf{59.30\%}       & \textbf{72.20\% }         & \textbf{70.29\% }& \textbf{70.38\%} \\ \bottomrule
\end{tabular}
\end{table} \textbf{Ablation experiments.} 
To assess the effectiveness of Activation-Weight Equalization (AWE) and Bias Correction(BC), we conducted quantization ablation experiments with 8-bit weights and 8-bit activations (W8A8) on the OPT-175B model. The experimental results are presented in Table \ref{tab:ablation}. 
We observed that only adding the BC operation to the W8A8 quantization model did not significantly improve model performance. In fact, for simple tasks like ARC-e, there was even a decrease in accuracy. However, only W8A8+AWE quantization led to a substantial increase in accuracy. The best results were obtained when both BC and AWE were used together. Our conclusion is that the BC operation can bring greater improvements to quantized models after the application of the AWE operation. 

\begin{table}[!ht]
\centering
\caption{Ablation experiments were conducted on the OPT-175B model. \textbf{W8A8+BC} represents the addition of the BC operation exclusively on the W8A8 base model.\textbf{ W8A8+AWE} denotes the inclusion of the AWE operation solely on the W8A8 model. \textbf{AWEQ} stands for the combination of both AWE and BC operations on the W8A8 base. The best experimental results are highlighted in bold.
}
\label{tab:my-table}
\begin{tabular}{ccc|cccc}
\toprule
   \textbf{OPT-175B}  & AWE & BC & PIQA  & HellaSwag & WinoGrande & ARC-e  \\ \midrule
   FP16 &-&-& 79.76\% &59.37\%   &        72.68\%  &71.10\% \\ \midrule
   W8A8 &\XSolidBrush&\XSolidBrush &      53.43\%  &25.61\%  &50.32\% & 48.26\%  \\
   W8A8+BC &\XSolidBrush&\CheckmarkBold&  53.45\%  &25.61\%  &50.36\% & 48.22\%  \\
   W8A8+AWE &\CheckmarkBold&\XSolidBrush& 79.69\%  &59.28\%  &72.11\% & 70.02\%  \\
   AWEQ &\CheckmarkBold&\CheckmarkBold&  \textbf{ 79.72\%}  &\textbf{59.30\%}  &\textbf{72.20\%} & \textbf{70.29\%} \\  \bottomrule
\end{tabular}
\label{tab:ablation}
\end{table}
\section{Conclusion} 
We introduce a post-training quantization approach, called AWEQ, which achieves state-of-the-art results in both ultra-low-bit quantization and INT8 quantization. AWEQ does not require additional training overhead. Moreover, this method eliminates the need for complex second-order matrix operations and hardware-unfriendly mixed-precision quantization like GPTQ. By equalizing activations and weights to the same range, it reduces wasted quantization grid points caused by outliers, thus maximizing the preservation of the original model's information. Furthermore, we introduce Bias Correction(BC) for mitigating the bias errors caused by quantization in LLMs. Extensive experiments were conducted on the LLaMA family and OPT-175B, and the results demonstrate that AWEQ outperforms existing LLMs quantization methods. 
\Section{References}
\bibliographystyle{IEEEbib}
\bibliography{refs}

\end{document}